\documentclass[11pt]{article}

\usepackage{acl}
\usepackage{fontspec}

\usepackage{times}
\usepackage{latexsym}
\usepackage[T1]{fontenc}
\usepackage[utf8]{inputenc}
\usepackage{microtype}
\usepackage{graphicx}
\usepackage{booktabs}
\usepackage{amsmath}
\usepackage{multirow}
\usepackage{tabularx}
\usepackage{placeins}
\usepackage{url}

\title{A Benchmark for Bangla Task-Oriented Dialogue Understanding}

\author{Afrina Alam Lamia\thanks{These authors contributed equally to this work.} \\
  East West University \\ 
  Dhaka, Bangladesh \\ \And
  Sumiaya Sharif\footnotemark[1] \\
  East West University \\ 
  Dhaka, Bangladesh \\ \And
  Prottoy Modak\footnotemark[1]\\
  East West University \\ 
  Dhaka, Bangladesh \\ \AND
  Ohiduzzaman Rayhan\footnotemark[1]\\
  East West University \\ 
  Dhaka, Bangladesh \\ \And
  Nafis Sadeq \\
  East West University \\ 
  Dhaka, Bangladesh}

\begin{document}
\maketitle

\begin{abstract}
Task-oriented dialogue understanding remains underexplored for Bangla, particularly in multilingual settings involving English and romanized Bangla (Banglish). We introduce \textbf{BanglaTOD}, a parallel tri-lingual benchmark comprising 209 multi-turn dialogues across nine Bangladesh-specific service domains and 64 intents. Each conversational turn is aligned across Bangla, English, and Banglish, yielding 4,689 language-specific turn records annotated with intents, token-level slots, state updates, and accumulated dialogue states. We establish context-aware baselines for joint intent classification and slot filling using mBERT, BERT-base-uncased, and BanglaBERT. Cross-language analysis reveals substantial performance differences across models and linguistic representations, while removing dialogue history reduces intent accuracy by 19.0--25.7 percentage points. BanglaTOD provides a foundation for multilingual dialogue understanding and future dialogue state tracking research in Bangladesh-specific conversational applications. The dataset is publicly available at \href{https://huggingface.co/datasets/afri7here/BanglaTOD}{\texttt{Hugging Face}}.
\end{abstract}

\section{Introduction}
\label{sec:intro}
Task-oriented dialogue systems rely on identifying user intents, extracting relevant slot values, and maintaining conversational context across turns~\cite{budzianowski-etal-2018-multiwoz,chen2019bert,10.1162/tacl_a_00314,rastogi2020towards}. Despite advances in multilingual natural language understanding~\cite{schuster2019cross,ding-etal-2022-globalwoz}, resources for Bangla task-oriented dialogue remain limited~\cite{khan2021nlu,karim-uzuner-2025-leveraging,sadeq2020icassp}. This limitation is particularly relevant in Bangladesh, where digital communication frequently involves Bangla, English, and Banglish~\cite{alam-etal-2025-bnsentmix,hossain2026jobccc,bari2026daraz}. Variations in language, script, and spelling introduce additional challenges for intent recognition, slot extraction, and dialogue state tracking.

Existing multilingual resources, including MultiATIS++~\cite{xu-etal-2020-end}, MTOP~\cite{li-etal-2021-mtop}, and GlobalWoZ~\cite{ding-etal-2022-globalwoz}, have advanced cross-lingual task-oriented NLU but provide limited coverage of Bangladesh-specific conversational settings. Bangla NLP resources have primarily addressed language modeling, general language understanding, and generation~\cite{bhattacharjee2022banglabert,bhattacharjee-etal-2023-banglanlg}. More recently, Bangla intent classification and slot filling have been investigated using translated task-oriented datasets~\cite{karim-uzuner-2025-leveraging}. However, resources containing multi-turn conversations with parallel Bangla, English, and Banglish utterances and structured dialogue-state annotations remain scarce. Translated corpora also do not contain realistic ontology for Bangladesh.

To address this gap, we introduce \textbf{BanglaTOD}, a parallel tri-lingual benchmark for task-oriented dialogue understanding in Bangladesh. The dataset comprises 209 manually constructed multi-turn dialogues spanning nine service domains and 64 intents. Each conversational turn is aligned across Bangla, English, and Banglish, producing 4,689 language-specific turn records. The annotations include intents, token-level slots, turn-level state updates, and accumulated dialogue states, supporting both utterance-level understanding and future dialogue state tracking research. We additionally establish context-aware baselines using three pretrained encoders and examine their performance across linguistic representations.

Our main contributions are:
\begin{itemize}
    \item \textbf{A turn-aligned multilingual dialogue resource:}
    We introduce BanglaTOD, comprising 209 dialogues across nine Bangladeshi service domains and 64 intents, with parallel Bangla, English, and Banglish utterances annotated for intent, slots, and dialogue states.

    \item \textbf{Context-aware multilingual benchmarks:}
    We evaluate mBERT, BERT-base-uncased, and BanglaBERT for joint intent classification and slot filling across all three linguistic representations.

    \item \textbf{Analysis of language and context:}
    We examine cross-language performance differences, the contribution of dialogue history, and recurring intent and slot prediction errors, highlighting challenges for multilingual task-oriented dialogue understanding.
\end{itemize}

\section{Dataset and Methodology}
\label{sec:dataset}

\subsection{Dataset Coverage and Construction}
BanglaTOD comprises 209 multi-turn dialogues across nine Bangladesh-specific service domains and 64 intents. Domains were selected for their relevance to everyday service interactions, informed by existing task-oriented dialogue resources. Four annotators constructed the dialogues through pairwise role-playing (user and system), to simulate realistic service interactions. Each intent represents a distinct user goal within its domain. All dialogues and annotations were manually reviewed to identify and correct errors. Table~\ref{tab:domains} summarizes the dataset coverage.

Dialogues were constructed as user--system interactions spanning 2--10 turns. They include follow-up requests that refer to information introduced in earlier turns, capturing contextual dependencies beyond isolated utterances.

\begin{table}[h]
\centering
\small
\begin{tabular}{@{}lrr@{}}
\toprule
\textbf{Domain} & \textbf{Dialogues} & \textbf{Intents} \\
\midrule
Education             & 50  & 9  \\
Government Service    & 32  & 19 \\
Healthcare             & 25  & 5  \\
Mobile Financial Serv. & 25  & 5  \\
E-commerce             & 25  & 10 \\
Restaurant             & 16  & 5  \\
Tourist                & 14  & 5  \\
Travel                 & 12  & 4  \\
Hotel                  & 10  & 2  \\
\midrule
\textbf{Total} & \textbf{209} & \textbf{64} \\
\bottomrule
\end{tabular}
\caption{Domain and intent coverage of the dataset.}
\label{tab:domains}
\end{table}

\subsection{Parallel Representations and Annotations}

Each dialogue is represented in Bangla, English, and Banglish, with corresponding conversational turns aligned across all three representations. The shared dialogue structure enables controlled comparisons of equivalent conversational content across languages and scripts. Each dialogue contains intent labels and structured slot annotations. A slot annotation specifies its type, textual expression, canonical value, and token boundaries. To capture information accumulated across turns, state update records newly introduced or modified values, while dialogue state stores the accumulated state after each turn.

The dataset is distributed in JSON format. Each dialogue contains a unique identifier, domain and intent information, and an ordered sequence of turns. Each turn records its speaker and the three aligned annotated utterance representations. This structure supports intent classification, slot filling, and dialogue state tracking.

\subsection{Dataset Statistics}

BanglaTOD contains 1,563 conversational turns, represented as 4,689 language-specific turn records. Dialogues average 7.48 turns, and the dataset contains 348 unique slot types. Table~\ref{tab:lexical_statistics} summarizes lexical differences across representations. English utterances have the highest average token count, while Banglish exhibits the largest vocabulary. 

\begin{table}
\small
\centering
\begin{tabular}{lr}
\hline
\textbf{Characteristic} & \textbf{Value} \\
\hline
Total dialogues & 209 \\
Languages & 3 \\
Domains & 9 \\
Intents & 64 \\
Turns per language & 1,563 \\
Language-specific turn records & 4,689 \\
Average turns per dialogue & 7.48 \\
Minimum turns per dialogue & 2 \\
Maximum turns per dialogue & 10 \\
Unique slot types & 348 \\
\hline
\end{tabular}
\caption{Overview of BanglaTOD.}
\label{tab:dataset_overview}
\end{table}

\begin{table}
\small
\centering
\begin{tabular}{lrrr}
\hline
\textbf{Language} & \textbf{Tokens} & \textbf{Vocabulary} & 
\textbf{Avg. Tokens/Turn} \\
\hline
Bangla & 15,726 & 2,448 & 10.06 \\
English & 17,858 & 2,192 & 11.43 \\
Banglish & 15,412 & 2,661 & 9.86 \\
\hline
\end{tabular}
\caption{Lexical statistics for the language variants.}
\label{tab:lexical_statistics}
\end{table}

\section{Experimental Setup}

\paragraph{Data and Split.}
We partition the 209 dialogues into training (137), validation (36), and test (36) sets using dialogue-level split with intent-based stratification. All three parallel representations of each dialogue remain in the same partition to prevent cross-language leakage.

\paragraph{Models and Training.}
We fine-tune three encoders:  bert-base-multilingual-cased (mBERT)~\cite{devlin2019bert}, BERT-base-uncased (BERT-base)~\cite{devlin2019bert}, and sagorsarker/bangla-bert-base (BanglaBERT)~\cite{Sagor_2020}. Each encoder is trained on pooled Bangla, English, and Banglish data and evaluated separately on each language. Following~\citet{chen2019bert}, the joint architecture uses a two-layer intent classifier (768$\to$256$\to$64) and a linear BIO slot classifier. Domain predictions are derived from the predicted intents. Models are trained using AdamW with a weighted joint loss ($\mathcal{L}_{int}+1.5\mathcal{L}_{slot}$) across three seeds (42, 43, 44). Additional training details are provided in Appendix~\ref{app:training}. Each target utterance is encoded with up to three preceding turns from the same dialogue and language, separated by \texttt{[SEP]} tokens~\citep{padmakumar-etal-2023-investigating}. %

\paragraph{Evaluation.}
We report intent accuracy (equivalent to micro-F1 for single-label classification), macro-F1, and weighted-F1. Slot filling is evaluated using entity-level precision, recall, and F1 based on exact span and type matches. We additionally report domain accuracy and joint exact match (EM), which requires both a correct intent and an exact match of the complete slot-label sequence. Results are averaged across three seeds.

\begin{table*}[t]
\centering
\setlength{\tabcolsep}{4pt}
\renewcommand{\arraystretch}{1.1}
\footnotesize
\begin{tabular}{lcccccccc}
\toprule
& \multicolumn{4}{c}{Intent / Domain} & \multicolumn{3}{c}{Slot (entity-level)} & Joint\\
\cmidrule(lr){2-5}\cmidrule(lr){6-8}\cmidrule(lr){9-9}
Model / Language & Dom.\ Acc & Micro-F1 & Macro-F1 & Wtd-F1 & Micro-F1 & Macro-F1 & Wtd-F1 & EM\\
\midrule
\multicolumn{9}{l}{\textbf{mBERT}}\\
\hspace{1em}Overall  & 0.859$_{\pm.007}$ & 0.601$_{\pm.012}$ & 0.408$_{\pm.015}$ & 0.623$_{\pm.012}$ & 0.330$_{\pm.003}$ & 0.194$_{\pm.003}$ & 0.302$_{\pm.006}$ & 0.219$_{\pm.010}$\\
\hspace{1em}Bangla   & 0.869$_{\pm.018}$ & 0.605$_{\pm.019}$ & 0.453$_{\pm.007}$ & 0.613$_{\pm.019}$ & 0.323$_{\pm.011}$ & 0.203$_{\pm.003}$ & 0.285$_{\pm.015}$ & 0.225$_{\pm.009}$\\
\hspace{1em}English  & 0.859$_{\pm.006}$ & 0.625$_{\pm.007}$ & 0.482$_{\pm.009}$ & 0.636$_{\pm.010}$ & 0.345$_{\pm.006}$ & 0.231$_{\pm.009}$ & 0.318$_{\pm.004}$ & 0.237$_{\pm.006}$\\
\hspace{1em}Banglish & 0.849$_{\pm.007}$ & 0.573$_{\pm.013}$ & 0.426$_{\pm.014}$ & 0.592$_{\pm.008}$ & 0.321$_{\pm.005}$ & 0.212$_{\pm.004}$ & 0.301$_{\pm.004}$ & 0.195$_{\pm.018}$\\
\midrule
\multicolumn{9}{l}{\textbf{BERT-base}}\\
\hspace{1em}Overall  & 0.822$_{\pm.011}$ & 0.551$_{\pm.034}$ & 0.348$_{\pm.031}$ & 0.565$_{\pm.037}$ & 0.285$_{\pm.007}$ & 0.182$_{\pm.003}$ & 0.255$_{\pm.006}$ & 0.189$_{\pm.001}$\\
\hspace{1em}Bangla   & 0.758$_{\pm.018}$ & 0.435$_{\pm.041}$ & 0.317$_{\pm.038}$ & 0.440$_{\pm.045}$ & 0.215$_{\pm.014}$ & 0.124$_{\pm.013}$ & 0.174$_{\pm.012}$ & 0.148$_{\pm.012}$\\
\hspace{1em}English  & 0.874$_{\pm.016}$ & 0.620$_{\pm.035}$ & 0.458$_{\pm.029}$ & 0.623$_{\pm.043}$ & 0.308$_{\pm.009}$ & 0.222$_{\pm.007}$ & 0.283$_{\pm.007}$ & 0.207$_{\pm.012}$\\
\hspace{1em}Banglish & 0.835$_{\pm.009}$ & 0.600$_{\pm.026}$ & 0.429$_{\pm.014}$ & 0.605$_{\pm.029}$ & 0.329$_{\pm.011}$ & 0.231$_{\pm.003}$ & 0.295$_{\pm.008}$ & 0.210$_{\pm.004}$\\
\midrule
\multicolumn{9}{l}{\textbf{BanglaBERT}}\\
\hspace{1em}Overall  & 0.675$_{\pm.011}$ & 0.380$_{\pm.017}$ & 0.230$_{\pm.016}$ & 0.393$_{\pm.023}$ & 0.192$_{\pm.007}$ & 0.099$_{\pm.006}$ & 0.165$_{\pm.005}$ & 0.123$_{\pm.007}$\\
\hspace{1em}Bangla   & 0.832$_{\pm.029}$ & 0.659$_{\pm.042}$ & 0.478$_{\pm.041}$ & 0.667$_{\pm.049}$ & 0.288$_{\pm.005}$ & 0.180$_{\pm.009}$ & 0.252$_{\pm.005}$ & 0.203$_{\pm.024}$\\
\hspace{1em}English  & 0.583$_{\pm.004}$ & 0.232$_{\pm.019}$ & 0.149$_{\pm.021}$ & 0.220$_{\pm.031}$ & 0.144$_{\pm.013}$ & 0.064$_{\pm.011}$ & 0.108$_{\pm.011}$ & 0.077$_{\pm.007}$\\
\hspace{1em}Banglish & 0.610$_{\pm.015}$ & 0.249$_{\pm.004}$ & 0.155$_{\pm.006}$ & 0.227$_{\pm.011}$ & 0.129$_{\pm.017}$ & 0.059$_{\pm.011}$ & 0.096$_{\pm.014}$ & 0.089$_{\pm.006}$\\
\bottomrule
\end{tabular}
\caption{Test results as mean$_{\pm\text{std}}$ over seeds 42, 43, 44 (context-aware input). Domain accuracy is derived from the predicted intent. Dom.\ = domain, Wtd = weighted, EM = exact match, Std = standard deviation.}
\label{tab:main}
\end{table*}

\section{Results and Discussion}

\subsection{Cross-Language Analysis}
\label{sec:cross-language}
Table~\ref{tab:main} shows substantial differences across the three language representations, despite their parallel conversational content. mBERT achieves the strongest overall performance and remains relatively consistent across languages, with intent accuracy ranging from 0.573 to 0.625. In contrast, BERT-base performs substantially better on English and Banglish than on Bangla, while BanglaBERT achieves its highest intent accuracy on Bangla (0.659) but transfers poorly to the other representations.

These patterns are consistent with differences in pretraining language and script coverage~\cite{rust-etal-2021-good, wu-dredze-2020-languages}. The parallel structure of BanglaTOD enables such comparisons while controlling for conversational content, revealing the challenges of supporting native-script, English, and romanized language representations within a shared dialogue system.

\subsection{Fine-Grained Intent and Slot Challenges}

Across all backbones, domain classification is considerably more accurate than intent classification. For example, mBERT achieves 0.859 domain accuracy
but only 0.601 intent accuracy. Intent macro-F1 is also consistently lower than weighted-F1 (0.408 versus 0.623 for mBERT), suggesting difficulties
with infrequent intent categories.

Error inspection further reveals confusion between semantically related intents, including cash\_out\_information and transaction\_support, and between tour\_cancellation and modify\_booking. Government-service intents, which constitute the largest intent inventory in the dataset, are particularly challenging. These observations highlight the difficulty of distinguishing fine-grained user goals in a limited-data setting.

Slot filling presents an additional challenge. Even the strongest model achieves only 0.330 entity-level F1. Errors frequently involve omitted slots and incomplete multi-word spans, whereas short, regularly formatted values are recovered more reliably. Together, these results suggest that fine-grained intent recognition and structured slot extraction remain important challenges for low-resource task-oriented dialogue understanding.

\subsection{Effect of Dialogue Context}

To assess the importance of conversational history, we retrain each backbone without preceding turns.
Table~\ref{tab:context} summarizes the ablation. Removing context reduces intent accuracy by 19.0--25.7 percentage points across the three backbones. The corresponding reductions in token-level slot macro-F1 are smaller. This pattern suggests that preceding turns are particularly useful for resolving context-dependent user intentions, such as follow-up questions that omit previously established information. The consistent improvements across backbones demonstrate the importance of multi-turn context and motivate evaluation beyond isolated utterances.

\begin{table}[t]
\centering\footnotesize
\setlength{\tabcolsep}{3pt}
\begin{tabular}{llccc}
\toprule
Model & Metric & Ctx & No Ctx & $\Delta$\\
\midrule
mBERT & Intent acc. & 0.605 & 0.356 & +0.249\\
 & Token macro-F1 & 0.250 & 0.215 & +0.035\\
\midrule
BERT-base. & Intent acc. & 0.578 & 0.321 & +0.257\\
 & Token macro-F1 & 0.216 & 0.179 & +0.037\\
\midrule
BanglaBERT & Intent acc. & 0.385 & 0.195 & +0.190\\
 & Token macro-F1 & 0.128 & 0.106 & +0.021\\
\bottomrule
\end{tabular}
\caption{Effect of dialogue context. Ctx = up to three previous turns; $\Delta$ = Ctx $-$ No ctx.}
\label{tab:context}
\end{table}

\subsection{Intent Error Analysis}
\label{sec:error-analysis}

Table~\ref{tab:perclass} illustrates variation in per-intent F1. While hotel\_booking\_cancellation is recognized reliably across backbones, several government-service intents perform poorly. The latter belong to the domain with the largest intent inventory, including semantically overlapping categories. Recurring confusions involve related user goals. For mBERT, cash\_out\_information is predicted as transaction\_support in three of four instances in each language representation. Similarly, BERT-base confuses tour\_cancellation with modify\_booking across all three representations. These errors suggest that semantic overlap between intent categories contributes to classification difficulty.

\begin{table}
\centering
\small
\resizebox{0.99\columnwidth}{!}{
\begin{tabular}{lccc}
\toprule
Intent & mBERT & BERT & BanglaBERT\\
\midrule
\texttt{hotel\_book\_cancel}
    & 1.000 & 1.000 & 0.882\\
\texttt{nid\_card\_service}
    & 0.250 & 0.154 & 0.000\\
\texttt{police\_services}
    & 0.000 & 0.154 & 0.267\\
\texttt{govt\_fin\_assist}
    & 0.333 & 0.231 & 0.100\\
\texttt{land\_records}
    & 0.588 & 0.286 & 0.190\\
\texttt{tax\_services}
    & 0.000 & 0.476 & 0.286\\
\bottomrule
\end{tabular}
}
\caption{Selected per-intent F1 scores on the
test set at the seed-42 checkpoint.}
\label{tab:perclass}
\end{table}

Detailed results on slot errors appear in Appendix~\ref{app:slot-error-analysis}. Short, structured values such as booking\_id and date\_of\_birth are identified correctly, although their support is small. Other labels, such as application\_status, are omitted entirely. Generic labels including query and service show low precision, indicating difficulty distinguishing broad slot categories. Multi-word expressions present an additional challenge. The continuation label I-requested\_information achieves only 0.277 recall for mBERT. Similar omissions occur for the other backbones: BanglaBERT and BERT-base predict \texttt{O} for 47 and 30 of 65 requested\_information tokens, respectively. These errors help explain the low entity-level F1, where recovering an incomplete span does not count as a correct entity.

\section{Conclusion}

We introduce BanglaTOD, a parallel Bangla, English, and Banglish benchmark for multi-turn task-oriented dialogue understanding in Bangladesh-specific service domains. Baseline evaluations reveal substantial variation across language representations, persistent challenges in fine-grained intent classification and slot extraction, and clear benefits from incorporating dialogue history. These findings establish BanglaTOD as a resource for studying multilingual and context-dependent dialogue understanding in low-resource settings. Future work includes hierarchical intent modelling, Banglish normalization, and end-to-end dialogue state tracking.

\section*{Limitations}

BanglaTOD is a relatively small, manually constructed dataset with uneven coverage across domains, intents, and slot types. Role-played conversations may not fully capture the linguistic diversity, informal expressions, and spelling variations encountered in real-world interactions, particularly in Banglish. Although the parallel representations enable controlled comparisons, they do not fully reflect naturally occurring code-switching or language-specific conversational patterns.

\bibliography{references}

\appendix
\section{Additional Training Details}
\label{app:training}

All three backbones use identical optimization settings,
summarized in Table~\ref{tab:config}.

\begin{table}[ht]
\centering
\small
\begin{tabular}{@{}lp{0.62\columnwidth}@{}}
\toprule
\textbf{Parameter} & \textbf{Value}\\
\midrule
Optimizer & AdamW \citep{loshchilov2018decoupled}\\
Learning rate & $1.5\times10^{-5}$\\
LLRD & 0.95\\
Weight decay & 0.01\\
Scheduler & Cosine, 10\% warm-up\\
Batch size & 16\\
Epochs & $\leq100$, patience 8\\
Dropout & 0.15\\
Gradient clipping & 1.0\\
Loss & $\mathcal{L}_{int}+1.5\mathcal{L}_{slot}$\\
Intent loss & Class-weighted CE,
              label smoothing 0.1\\
Model selection & $0.5\,\mathrm{Acc}_{int}
              +0.5\,\mathrm{F1}_{slot}^{token,macro}$\\
Seeds & 42, 43, 44\\
\bottomrule
\end{tabular}
\caption{Shared training configuration.}
\label{tab:config}
\end{table}

For each target utterance $u_t$, the input concatenates up to three preceding turns $c_1,c_2,c_3$:

\[
[\mathrm{CLS}]\;u_t[\mathrm{SEP}]\; c_1\ c_2\ c_3\ [\mathrm{SEP}].
\]

The maximum input length is 256 subtokens, with no truncation observed in the training or validation examples. Slot supervision is restricted to the first subtoken of each word in the target utterance; context, special tokens, and continuation subtokens are masked from the slot loss.

\section{Slot Error Analysis}
\label{app:slot-error-analysis}

Table~\ref{tab:slot} summarizes representative token-level slot predictions for mBERT.

\begin{table}[ht]
\centering
\small
\setlength{\tabcolsep}{2.5pt}
\resizebox{\columnwidth}{!}{%
\begin{tabular}{llccc}
\toprule
Error type & Slot label & P & R & Support\\
\midrule
Reliable
    & \texttt{B-booking\_id}
    & 1.000 & 1.000 & 9\\
    & \texttt{B-date\_of\_birth}
    & 1.000 & 1.000 & 3\\
\midrule
Omitted
    & \texttt{B-action}
    & 1.000 & 0.500 & 6\\
    & \texttt{B-hotel\_name}
    & 1.000 & 0.333 & 3\\
    & \texttt{B-appl\_status}
    & 0.000 & 0.000 & 3\\
\midrule
Low precision
    & \texttt{B-query}
    & 0.208 & 0.333 & 15\\
    & \texttt{B-service}
    & 0.276 & 0.667 & 12\\
\midrule
Multi-word
    & \texttt{I-requ\_info}
    & 0.448 & 0.277 & 47\\
    & \texttt{I-query}
    & 0.857 & 0.286 & 21\\
\bottomrule
\end{tabular}%
}
\caption{Representative token-level slot predictions for mBERT at the seed-42 checkpoint. P = precision; R = recall; support = number of tokens.}
\label{tab:slot}
\end{table}

\end{document}